\documentclass{article}

\usepackage{character_a_movie_arxiv}

\usepackage[utf8]{inputenc} 
\usepackage[T1]{fontenc}    
\usepackage{hyperref}       
\usepackage{url}            
\usepackage{booktabs}       
\usepackage{amsfonts}       
\usepackage{nicefrac}       
\usepackage{microtype}      
\usepackage{xcolor}         
\usepackage{algorithm}
\usepackage{epsfig}
\usepackage{amsmath}
\usepackage{amssymb}
\usepackage{booktabs}
\usepackage{bm}
\usepackage{multirow}
\usepackage{algorithmicx}
\usepackage{algpseudocode}
\usepackage{makecell}
\usepackage{textcomp}
\usepackage{bbding}
\usepackage{microtype}
\usepackage{array}
\usepackage{setspace}
\usepackage{threeparttable}
\usepackage{diagbox}
\usepackage{wrapfig}
\usepackage{float}
\usepackage{bbm}
\usepackage{url}
\usepackage{mathdots}

\title{MovieCharacter: A Tuning-Free Framework for Controllable Character Video Synthesis}

\author{%
  David S.~Hippocampus\thanks{Use footnote for providing further information
    funding agencies.} \\
  Department of Computer Science\\
  Cranberry-Lemon University\\
  Pittsburgh, PA 15213 \\
  \texttt{hippo@cs.cranberry-lemon.edu} \\
}

\begin{document}

\maketitle
\vspace{-0.5cm}
\begin{figure}[H]
  \includegraphics[width=\textwidth]{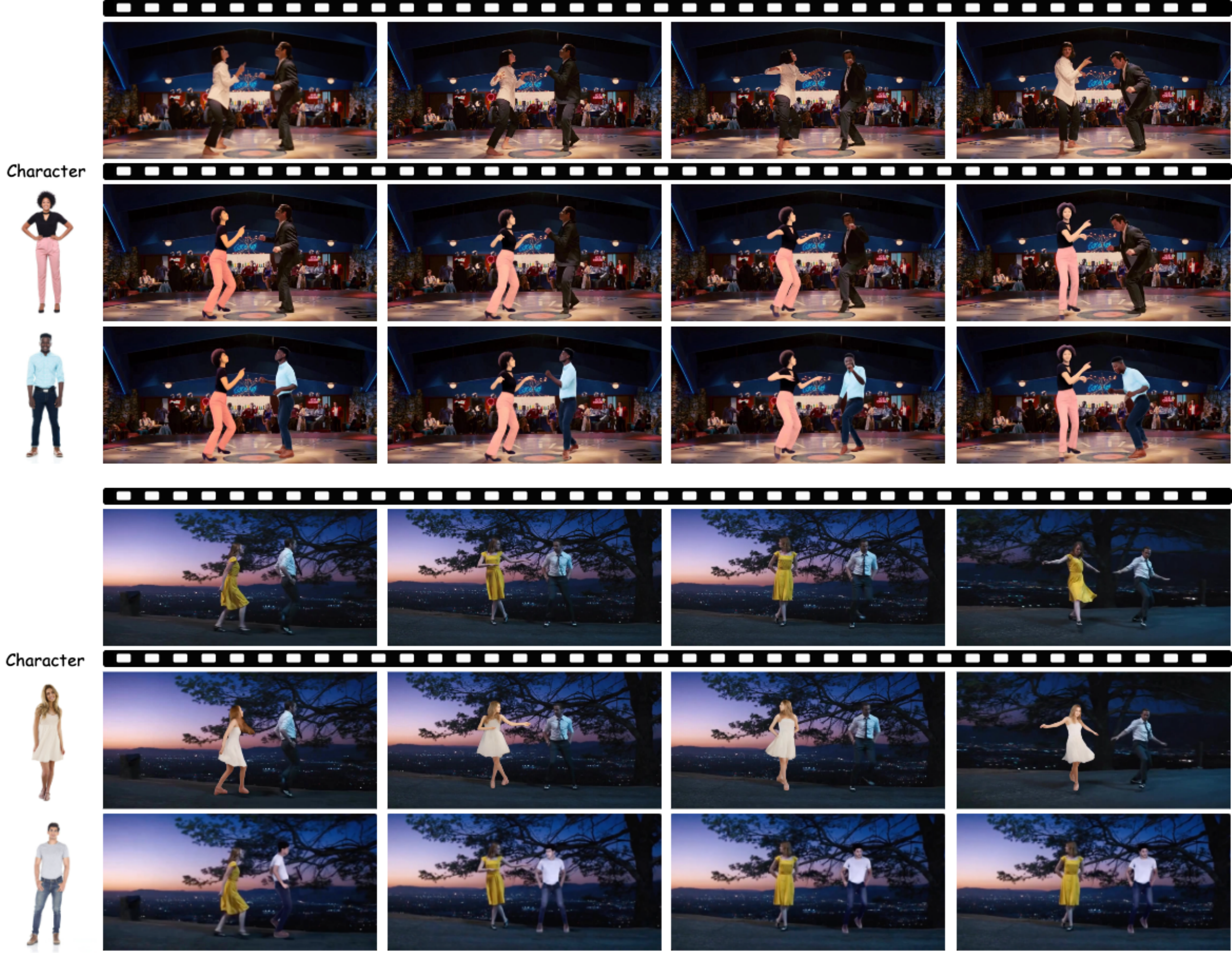}
  \caption{\textit{MovieCharacter} enables the replacement of any character in a movie with a 2D reference character, facilitating the synthesis of customized animated avatars. By utilizing driving motions sourced from movies, \textit{MovieCharacter} can generate movements that closely mimic the original character's actions. Additionally, our framework allows for the seamless integration of cinematic scenes, ensuring natural interactions between the synthesized characters and their environments. }
  \label{Fig: fig1}
\end{figure}

\begin{abstract}
Recent advancements in character video synthesis still depend on extensive fine-tuning or complex 3D modeling processes, which can restrict accessibility and hinder real-time applicability. To address these challenges, we propose a simple yet effective tuning-free framework for character video synthesis, named \textit{MovieCharacter}, designed to streamline the synthesis process while ensuring high-quality outcomes. Our framework decomposes the synthesis task into distinct, manageable modules: character segmentation and tracking, video object removal, character motion imitation, and video composition. This modular design not only facilitates flexible customization but also ensures that each component operates collaboratively to effectively meet user needs. By leveraging existing open-source models and integrating well-established techniques, \textit{MovieCharacter} achieves impressive synthesis results without necessitating substantial resources or proprietary datasets. Experimental results demonstrate that our framework enhances the efficiency, accessibility, and adaptability of character video synthesis, paving the way for broader creative and interactive applications.

\end{abstract}



\section{Introduction}

Character video synthesis has emerged as a critical challenge in the fields of computer vision and graphics, driven by its diverse applications in areas such as film production, video game development, virtual reality, and interactive media experiences. Recent advancements~\cite{zhong2024deco, men2024mimo,huang2024dreamwaltz,he2024id,fang2024motioncharacter,fei2024video} in this domain, such as neural rendering techniques and deep-generative models, have demonstrated promising results in producing realistic character animations and lifelike scenes. For instance, methods based on neural radiance fields (NeRF)~\cite{mildenhall2021nerf} and 3D Gaussian  splatting~\cite{kerbl20233d} have shown potential in capturing high-fidelity 3D representations of avatars, while approaches leveraging generative adversarial networks~\cite{goodfellow2020generative} and diffusion models~\cite{ho2020denoising, rombach2022high, song2020score} have advanced the generation of high-quality video content. However, many of the prevailing methods necessitate extensive fine-tuning or depend on complex 3D modeling techniques. These requirements not only hinder the accessibility of these approaches but also restrict their applicability in real-time scenarios, where efficiency and responsiveness are paramount. Consequently, there is an ongoing need for innovative solutions that can streamline the synthesis process, enabling the generation of high-quality character videos in a more efficient and user-friendly manner. Addressing these challenges will be essential for expanding the potential of character video synthesis in various creative and interactive applications.

In this paper, we introduce \textit{MovieCharacter}, a straightforward yet effective tuning-free framework designed for character video synthesis. This framework enables users to initiate synthesis with a selected movie and customize characters by decomposing the synthesis problem into distinct components, including character segmentation and tracking, video object removal, character motion imitation, and video composition. Each of these independent modules is designed to function collaboratively, allowing users to tailor the synthesis process according to their specific needs. Character segmentation and tracking facilitate the identification and customization of desired characters, while object removal enhances the visual coherence by eliminating unwanted elements. 
Character motion imitation leverages pose-driven techniques to capture and reproduce realistic movements, enabling the synthesized characters to mimic the nuanced dynamics and expressions of the original performance.
To achieve high-quality synthesis, our video composition module incorporates lighting-aware video harmonization, which adjusts the lighting of the target character to align with the ambient illumination of the background, thereby enhancing visual coherence. Additionally, edge-aware video refinement is utilized to smooth transitions between the character and the background, preserving fine details and improving the overall visual fidelity of the composite video. These components function in concert to provide visually coherent and temporally consistent character synthesis, resulting in compelling composite videos that maintain a high degree of fidelity to both the source characters and the target environment.


Particularly, our framework effectively leverages existing solutions and models from the open-source community, allowing for integration and adaptation of well-established techniques. Through thoughtful design and optimization, we can achieve impressive results without requiring extensive resources or proprietary datasets. \textit{MovieCharacter} enhances performance while promoting accessibility and flexibility, enabling developers to build on prior advancements in character video synthesis. In summary, our contributions are threefold:

\begin{itemize}
\item We introduce \textit{MovieCharacter}, a simple yet effective tuning-free framework for character video synthesis that decomposes the synthesis process into distinct modules, enabling nuanced control over individual aspects of the generated video while maintaining exceptional scalability, generalization capacity.

\item We incorporate a synergistic combination of lighting-aware and edge-aware video composition techniques, enhancing visual integration and maintaining fidelity to both source characters and the target environment, resulting in compelling and realistic composite videos.

\item We effectively leverage existing solutions and models from the open-source community, facilitating the integration of well-established techniques that enhance performance and accessibility, while promoting practical applications in character video synthesis.
\end{itemize}


\section{Related Work}

\subsection{Video Editing}
Video editing has undergone significant advancements in recent years, with diffusion models emerging as a powerful paradigm for high-quality manipulations. Diffusion models, initially proposed for image generation~\cite{ho2020denoising}, have recently been adapted for video editing tasks, offering unprecedented quality and flexibility. Recent developments have focused on enhancing the efficiency and applicability of diffusion-based video editing. Tune-A-Video~\cite{wu2023tune} introduced a method for fine-tuning pre-trained text-to-image diffusion models for text-driven video editing, significantly reducing computational requirements. Recently, a large number of works focus on zero-shot video editing~\cite{khachatryan2023text2video, qi2023fatezero, ceylan2023pix2video, yang2023rerender, bao2023latentwarp, cong2023flatten, geyer2023tokenflow, kara2024rave, yang2024fresco}, which leveraged pre-trained text-to-image diffusion models for video editing tasks without requiring additional training.

Despite these significant advancements, current video editing approaches, including those based on diffusion models, face limitations when it comes to character replacement with optional alternatives. Existing methods excel at tasks such as style transfer, object removal, or scene modification, but they lack the capability to seamlessly replace characters with user-selected alternatives while maintaining the original video's context, action, and temporal consistency. This limitation presents a significant challenge in scenarios where flexible character substitution is desired, such as in personalized content creation or adaptive storytelling applications. The development of a system capable of such selective character replacement while preserving video coherence remains an open challenge in the field of video editing.

\subsection{Character Video Synthesis}
Character video synthesis has gained significant attention in recent years, primarily due to the increasing demand for realistic animation and interactive applications. Traditional 3D methods often rely on multi-view captures and case-specific training, which not only limits their scalability but also hinders the ability to model arbitrary characters in a timely manner. Recent advancements~\cite{he2024id, fang2024motioncharacter} in 2D synthesis, particularly those leveraging pre-trained diffusion models, have made strides in overcoming some of these limitations. Depth information~\cite{feng2023dreamoving} and 3D human parametric models, such as SMPL~\cite{loper2023smpl, zhu2024champ}, can effectively represent human geometry and motion characteristics derived from source videos. Nevertheless, these dense guidance techniques may become overly dependent on signals from the source video, such as body outlines, This reliance can result in a degradation of the generated video quality, particularly when there is a significant discrepancy between the target identity and the source.

MIMO~\cite{men2024mimo} proposes a novel approach that addresses these shortcomings by allowing for the synthesis of character videos with controllable attributes, including character identity, motion dynamics, and interactive scenes-based solely on simple user inputs. However, it is important to note that the MIMO framework necessitates a substantial high-quality dataset and model training, which can be a non-trivial challenge for users to employ MIMO in their applications. In contrast, our proposed framework offers a tuning-free framework that achieves competitive results without the need for building high-quality datasets and training complex models, thereby promoting accessibility and efficiency in character video synthesis. 

\section{Methodology}

\begin{figure*}[t]
\begin{center}
\includegraphics[width=0.95\linewidth]{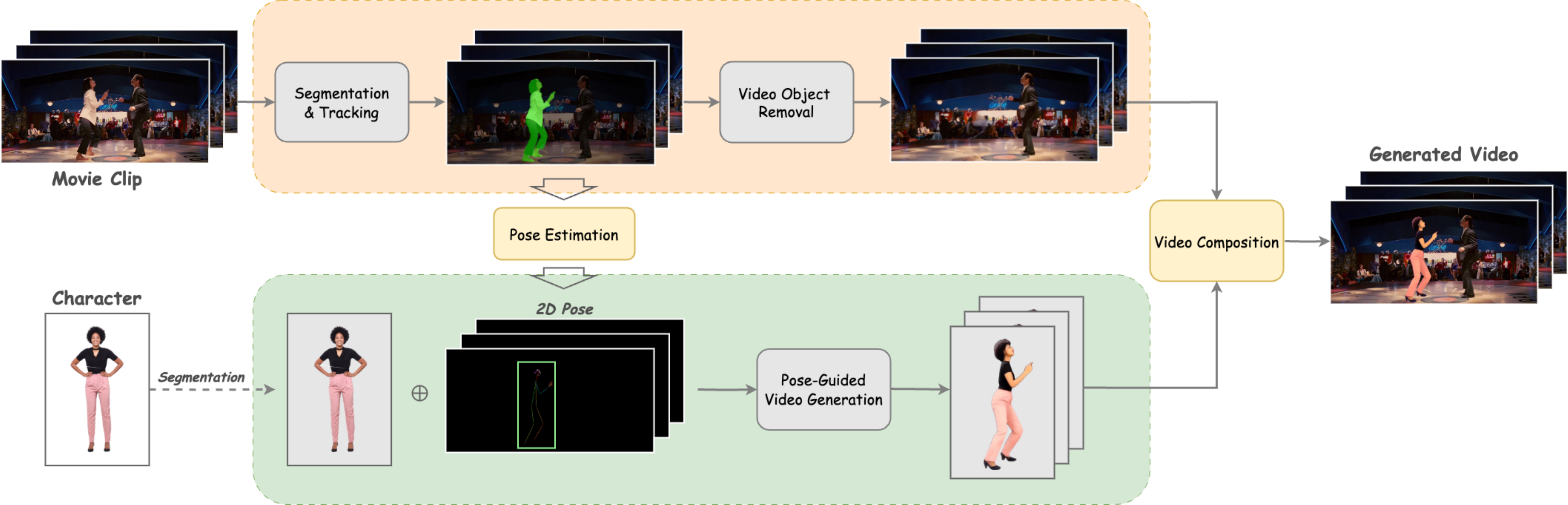}
\end{center}
   \caption{The overall architecture of \textit{MovieCharacter}, where the top part is the pipeline for detecting and removing the targeted character from the video, and the bottom part is character motion imitation branch. The composition module serves to integrate the outputs from the upper and lower branches of the framework.}
\label{fig1}
\end{figure*} 

\subsection{Problem Formulation}
In the problem of character replacement for movie scenes, our goal is to achieve character video synthesis by allowing the user to select the target subject through point-based interaction. Formally, given an input video sequence $\mathcal{V}$, a point prompt $\mathcal{P}$ indicating the subject to be replaced, and a character image $\mathcal{I}$ of the desired replacement characteristics, our approach generates a target synthesized video. 

The overview of the proposed framework is illustrated in Figure~\ref{fig1}, we decompose this task into several sub-problems, including enhancing the quality of local-aware character replacement, achieving character-controlled synthesis based on the given character image, and ensuring controllable replacement effects in terms of magnitude and appearance. 



\subsection{Character Segmentation and Tracking} 

\textit{MovieCharacter} requires precise segmentation of the character from the background, which can be initiated by user-provided spatial information. Users can provide the character's spatial information through various methods, such as clicking on specific points within the frame, defining a bounding box that encapsulates the character, or manually creating a mask that outlines the character's shape. These user inputs serve as critical prompts $\mathcal{P}$ for the segmentation model to accurately identify and isolate the character in the initial frame. To achieve a coherent replacement across the entire clip $\mathcal{V}$, the segmentation must be consistently tracked throughout all subsequent frames. In this paper, we employ the state-of-the-art Segment Anything 2 (SAM2)~\cite{ravi2024sam} to accomplish this task. SAM2 is a powerful tool that not only segments the character in the first frame but also tracks the segmentation across all frames, ensuring continuity and accuracy in the character replacement process.

The segmentation sequence obtained through SAM2 is crucial for downstream tasks. These include video object removal, where the selected character is completely removed from the scene, and 2D human pose estimation, which involves analyzing the character's posture and movement within the frame. The accuracy and robustness of the segmentation sequence significantly influence the effectiveness of these subsequent tasks, underscoring its critical role in the overall synthesis process.


\subsection{Video Object Removal}


The intuitive method to achieve character synthesis is by overlaying the pose-driven movements of the target character onto the original movie clip. This way is feasible based on the alignment of the target character's pose with that in the movie scene, ensuring a congruent integration. The underlying assumption is that the pose, as a set of abstract movement data, can be directly applied to the old clip without compromising the visual coherence. However, this simplicity may come at the cost of visual degradation. The direct pasting of poses may not account for the nuanced differences between the old and new characters, such as variations in body types, clothing shapes, and motion dynamics, which can lead to a noticeable disparity in the final output.

\begin{figure}[H]
  \includegraphics[width=\textwidth]{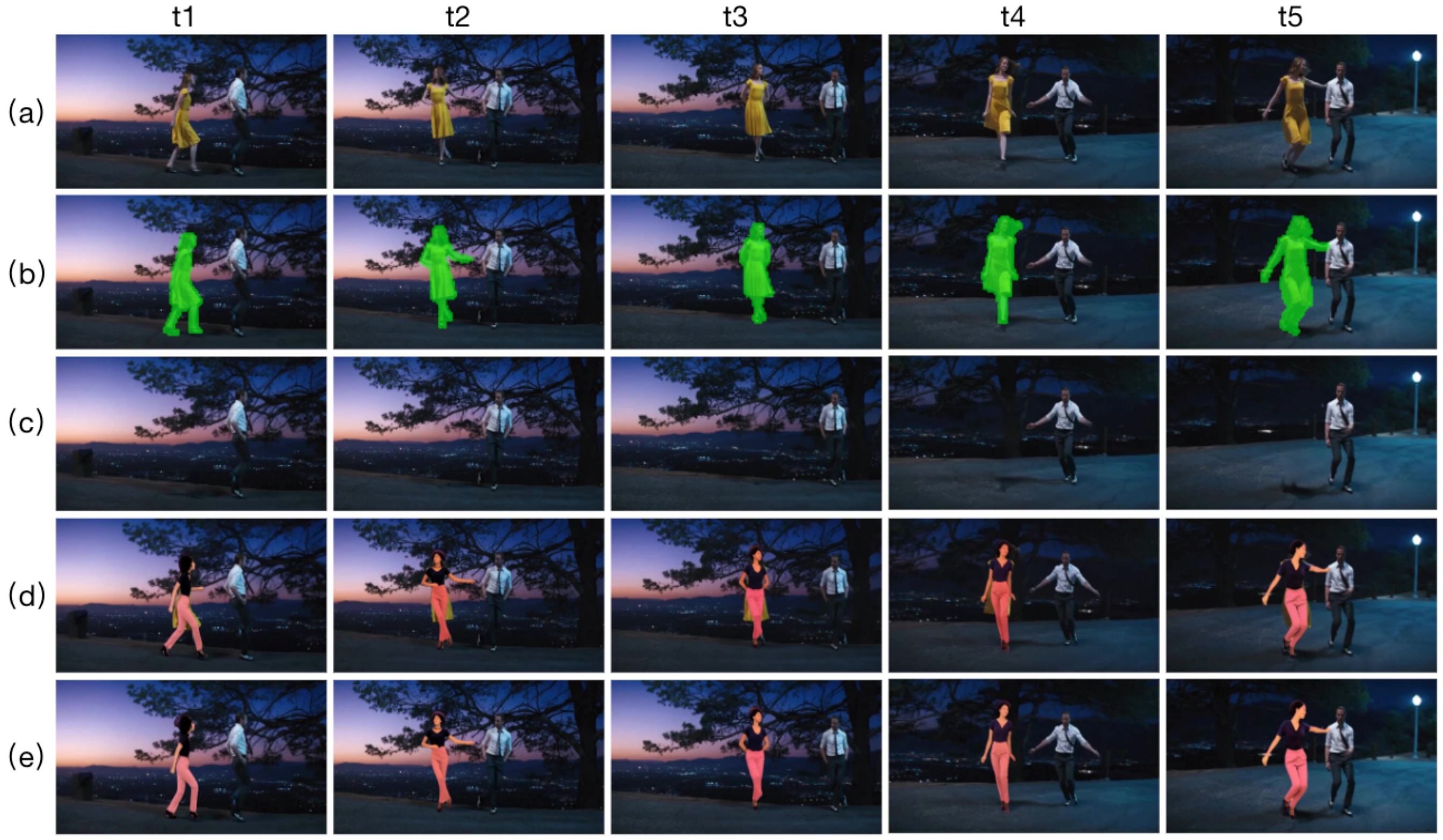}
  \caption{Video object removal is necessary for our proposed \textit{MovieCharacter}. The resultant frames at different time steps (column-wise) are presented with different operations (row-wise). (a) The original video clip. (b) The tracked segmentation (green color) of the object that is expected to be changed. (c) Video clip after applying video object removal. (d) Composition of the new character and the video clip without object removal. (e) Composition with object removal.}
  \label{fig:removal}
\end{figure}

To address these potential discrepancies and enhance the visual quality of the composited video, a more meticulous approach is used. This involves meticulously removing all traces of the old character from the footage, thereby providing a clean video background for the new character's integration. The process of erasing the old character and its associated elements is non-trivial and requires sophisticated tools to ensure that the background remains intact and free from any artifacts. To this end, we employ ProPainter~\cite{zhou2023propainter}, a state-of-the-art video inpainting method. This method is adept at filling in the gaps left by the removed character, ensuring a seamless and visually pleasing transition from the old to the new character. By leveraging ProPainter, we can achieve a high degree of realism in the final composite, aligning with the aesthetic standards expected in contemporary video production, as illustrated in Figure~\ref{fig:removal}.

\subsection{Character Motion Imitation}
Character motion imitation aims to enable a customized character $\mathcal{I}$ to replicate the movements of a target character from a selected movie, ensuring that the synthesized motions are consistent with the target's actions and perspectives. In this work, we reconceptualize the Character Motion Imitation task as a pose-guided character animation problem. Recent advancements in pose-guided character animation have largely focused on diffusion models, which effectively capture complex motion dynamics through high-dimensional pose representations. 

\noindent \textbf{Latent Diffusion Models.} 
Stable Diffusion~\cite{rombach2022high} is a prominent diffusion model that distinguishes itself by operating within the latent space. This innovative approach utilizes a pretrained autoencoder to perform two key functions: mapping images into latent representations and reconstructing these latents into high-resolution images. Starting from an initial input signal $z_0$, which is the latent of an input image \textit{$I_0$}, the process iteratively progresses through a series of steps. The diffusion forward process is defined as:
\begin{equation}
    q\left(z_t \mid z_{t-1}\right)=\mathcal{N}\left(z_t ; \sqrt{1-\beta_{t-1}} z_{t-1}, \beta_t I\right), \quad t=1, \ldots, T
\end{equation}
where $q\left(z_t \mid z_{t-1}\right)$ is the conditional density of $z_t$ given $z_{t-1}$, and $\beta_t$ is hyperparameters. $T$ is the total timestep of the diffusion process. The objective of the diffusion model is to learn the reverse process of diffusion, which is commonly known as denoising process. 
Specifically, the denoising process begins with the standard Gaussian distribution $p_\theta(z_{T})=\mathcal{N}(z_{T} ; 0, I)$ and iteratively removes the predicted noise at each step, until reaching the original signal $z_0$.
\begin{equation}
    p_\theta\left(z_{t-1} \mid z_t\right)=\mathcal{N}\left(z_{t-1} ; \mu_\theta\left(z_t, t\right), \Sigma_\theta\left(z_t, t\right)\right), \quad t=T, \ldots, 1
\end{equation}
where $\theta$ is the learnable parameters, trained for the inverse process. In Stable Diffusion, the model can be interpreted as a sequence of weight-sharing denoising autoencoders $\epsilon_\theta\left(z_t, t, c_{\mathcal{P}}\right)$, trained to predict the denoised variant of input $z_t$ and text prompt $c_{\mathcal{P}}$. The objective can be formulated as:
\begin{equation}
    \mathbb{E}_{z, \epsilon \sim \mathcal{N}(0,1), t}\left[\left\|\epsilon-\epsilon_\theta\left(z_t, t, c_{\mathcal{P}}\right)\right\|_2^2\right]
\end{equation}

\noindent \textbf{Pose-Guided Character Animation. }
The pose-guided video diffusion model is an mainstream approach designed to synthesize realistic video content by leveraging pose information as a guiding factor. This model operates by utilizing a diffusion process that progressively refines generated frames based on predefined pose parameters, enabling it to capture complex motion dynamics while adhering to the specified poses. By incorporating pose guidance, the model enhances its ability to produce coherent and fluid animations that are closely aligned with the intended character movements.

Several recent efforts~\cite{hu2024animate, wang2024unianimate, zhang2024mimicmotion, peng2024controlnext} have investigated the potential of leveraging diffusion models for image-guided character video synthesis, with promising results in mimicking complex movements and generating realistic animations.
In our framework, we build upon the Cogvideox~\cite{yang2024cogvideox} image-to-video diffusion model to develop a robust approach for generating high-quality, pose-guided human videos. By decomposing the motion representation into essential components, the Cogvideox-based approach allows for efficient capture and transfer of motion dynamics from the target character to the custom character. This decomposition enables our framework to preserve critical motion features while accommodating variations in appearance and body structure, thus achieving a balance between realism and adaptability. A comprehensive technical report detailing the implementation specifics and architectural design choices will be made publicly available in the near future.

Additionally, our framework is designed to be flexible, allowing for the integration of open-source techniques and models~\cite{hu2024animate, wang2024unianimate, zhang2024mimicmotion, peng2024controlnext} to achieve motion imitation capabilities. This flexibility not only simplifies implementation but also provides opportunities for incorporating state-of-the-art advancements from the open-source community, making it a practical and scalable solution for character motion imitation.

\subsection{Video Composition}
To achieve seamless integration of character motions, appearances, and scene elements into a visually cohesive composite output while maintaining high fidelity to both the source character attributes and environmental context, our framework introduces an advanced composition strategy that synergistically combines lighting-aware and edge-aware refinement mechanisms.

\noindent \textbf{Lighting-Aware Video harmonization. }
A major challenge in this process arises from discrepancies in illumination and shading. When a synthesized character is inserted into a modified scene, mismatches in lighting conditions between the original footage and the newly generated elements can become apparent, leading to unrealistic visual artifacts. These artifacts often manifest as misaligned shadows, inconsistent reflections, or unnatural highlights on the character, undermining the realism of the composition. 

We employ PCT-Net~\cite{guerreiro2023pct}, a method specifically designed for harmonizing the appearance of the composite foreground with the background. Recognizing that video frames are not independent but rather part of a continuous sequence, we process frames as contiguous blocks, ensuring temporal coherence within each block. Additionally, we smooth the transitions between these blocks to facilitate natural information flow and eliminate visual discontinuities. This approach minimizes inconsistencies, ensuring that the character's movements and interactions blend seamlessly within the scene, resulting in a realistic and harmonious final composition.

\begin{figure}[H]
  \includegraphics[width=\textwidth]{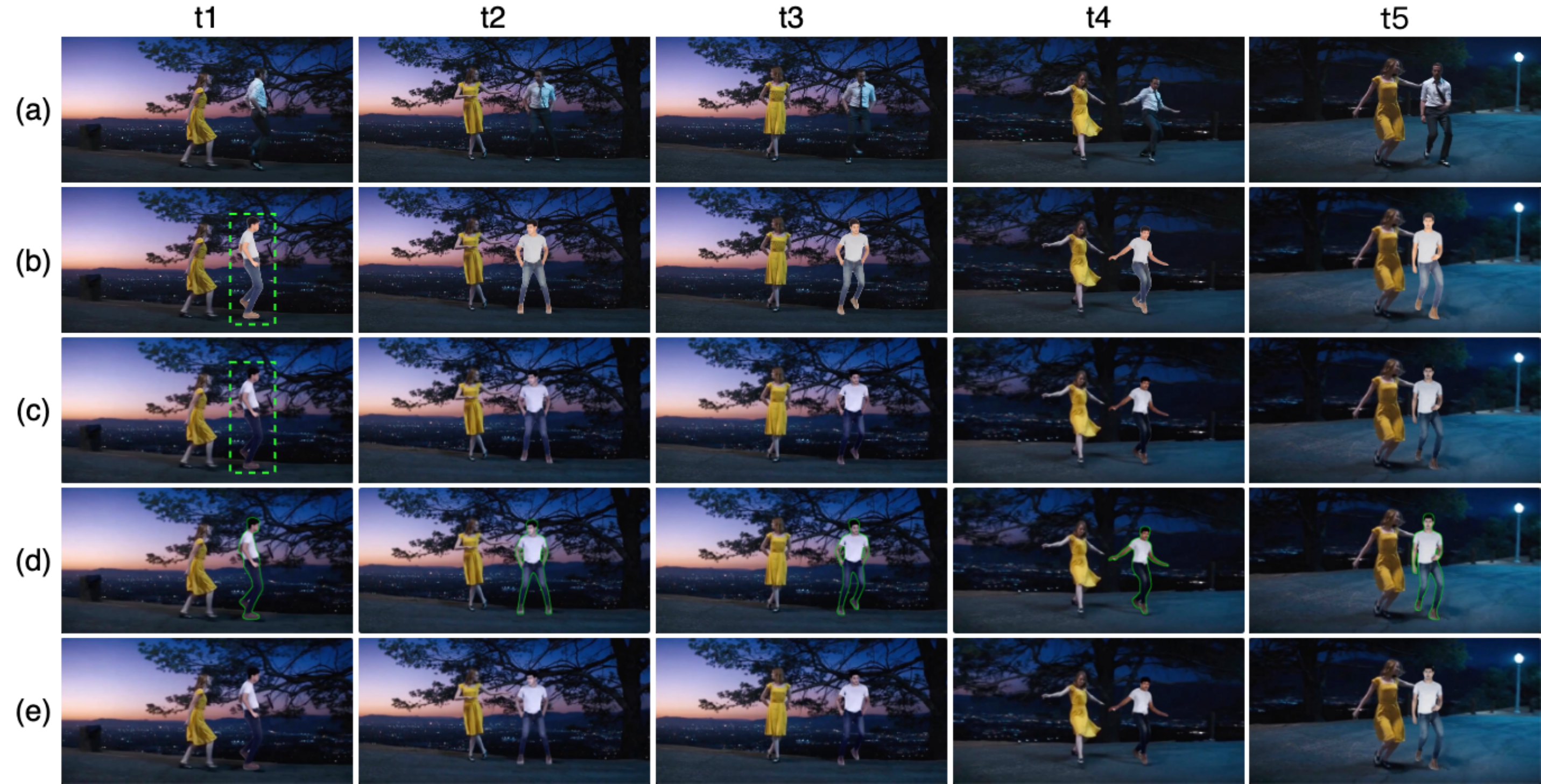}
  \caption{Compositing a character with a video clip requires further refinements. Frames at different time steps (column-wise) are displayed with different refinements (row-wise). (a) The original video clip. (b) Direct composition without refinements. A green dashed-line bounding box marks the new character. (c) Composition after video harmonization. (d) Composition overlapped with the segmentation (green color) of the edge area. (e) Composition with refined edges.}
  \label{fig:comp}
\end{figure} 

\noindent \textbf{Edge-Aware Video Refinement. }
While our video harmonization techniques effectively enhance the overall visual integration of the synthesized character with the background, some minor artifacts still persist along the edges of the character. These imperfections, though subtle, can detract from the overall visual quality of the final output and may disrupt the viewer's immersion. To address this challenge, we have reintroduced the ProPainter~\cite{zhou2023propainter} tool into our processing pipeline. This innovative approach diverges from traditional segmentation and tracking methods that focus solely on the object area; instead, we emphasize tracking the edge region itself. By honing in on the character's outline, we can more accurately capture its nuances, allowing for smoother transitions and improved edge fidelity in the composited video. This refinement enhances the visual coherence of the final output, contributing to a more polished and engaging viewing experience.

Subsequently, the mask sequence, which has undergone moderate dilation to ensure a smoother transition, is combined with the composited video. This pairing is then submitted to the ProPainter for further processing. The tool utilizes context-aware algorithms to fill in the edge areas with appropriate transitions, effectively mitigating the artifacts and enhancing the visual coherence of the character within the video scene. This refined method not only corrects the initial segmentation error but also improves the overall robustness of our video compositing technique.

\section{Experimental Results}

\begin{figure}[H]
  \includegraphics[width=\textwidth]{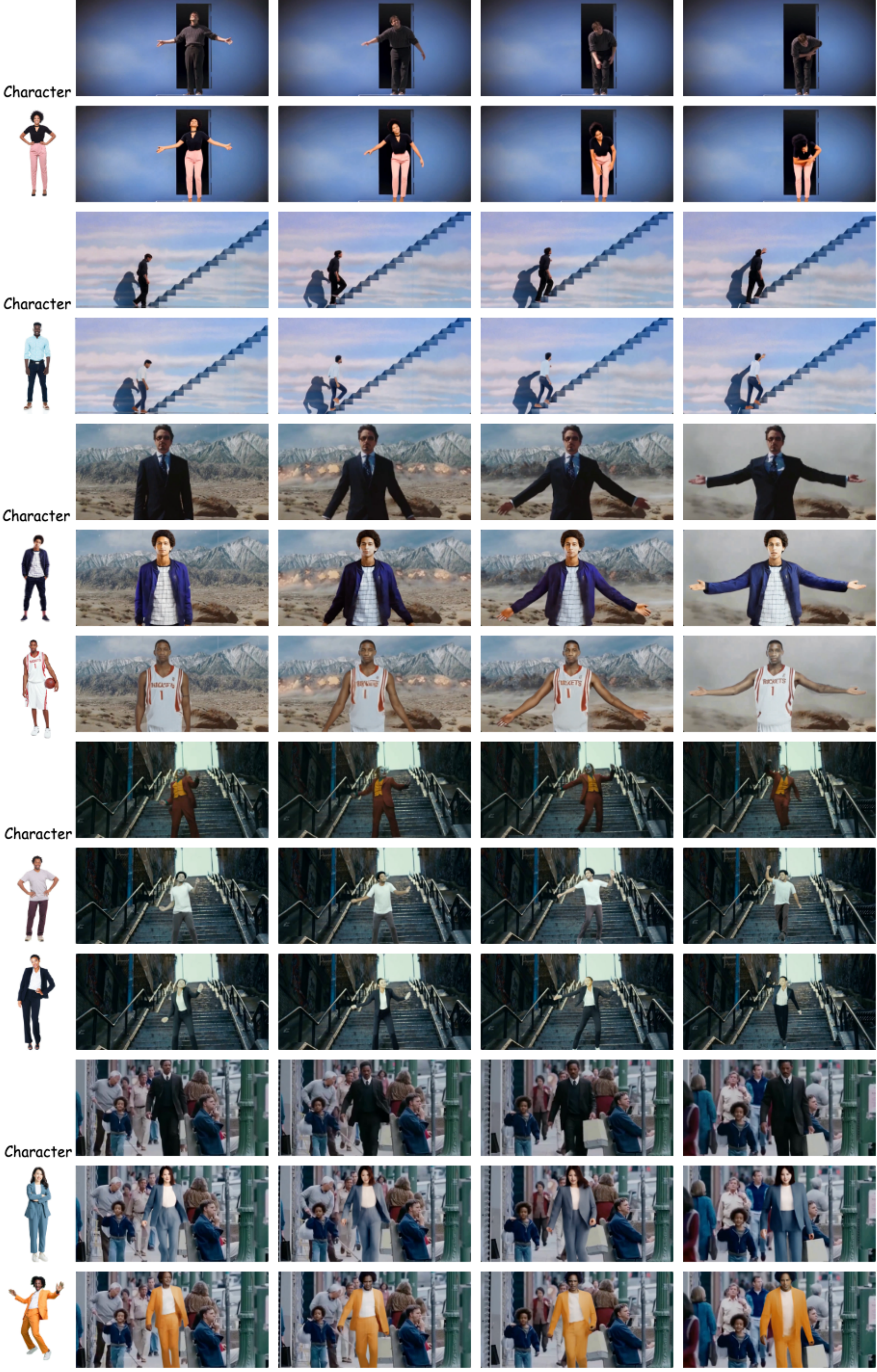}
  \caption{Examples of synthesizing avatar animations across multiple cinematic contexts, utilizing various reference characters.}
  \label{fig:exp}
\end{figure}

The character images used for testing are sourced from publicly available platforms, specifically \textcolor{magenta}{\url{https://www.pexels.com/}} and \textcolor{magenta}{\url{https://pixabay.com/}}. To evaluate the effectiveness of our proposed framework, we constructed a dataset comprising classic movie clips, which were collected from a widely used video-sharing platform\footnote{\textcolor{magenta}{\url{https://www.bilibili.com/}}}. In our experimental setup, the input reference images were resized to a resolution of 1024×768, while the input videos were configured at a resolution of 1024×2048. To thoroughly assess the robustness and generalization capability of our method, we performed an extensive series of experiments focused on character video synthesis. As illustrated in Figure~\ref{fig:exp}, the results demonstrate that our approach consistently produces high-quality outputs, with seamless integration of synthesized characters into the movie clips, validating the effectiveness of the proposed solution.

\noindent \textbf{Ablation Study.} 
In this study, we specifically compare the contributions of the video composition components, namely Lighting-Aware Video Harmonization and Edge-Aware Video Refinement. As shown in Figure~\ref{fig:comp}, we observe that the output depicted in Figure~\ref{fig:comp} (c) demonstrates improved character integration with the environment and overall consistency when compared to Figure~\ref{fig:comp} (b). Furthermore, the results shown in Figure~\ref{fig:comp} (e) reveal that the application of Edge-Aware Video Refinement leads to a noticeable enhancement in edge smoothness and visual fidelity compared to Figure~\ref{fig:comp} (c). These findings underscore the effectiveness of our proposed techniques in achieving high-quality character video synthesis.

\section{Limitation}
Despite achieving outstanding results, our methods still encounter several limitations. Our framework primarily stem from its reliance on the Character Motion Imitation module for character video synthesis in cinematic scenes. Currently, the framework struggles with representing complex action scenarios involving multiple interactions and occlusions effectively. While the overall pipeline established through this work enables the accumulation of data and the refinement of data preprocessing techniques, future efforts will focus on enhancing the capabilities of the Character Motion Imitation module. By addressing these limitations, we aim to improve the framework's performance in more intricate motion contexts, thereby expanding its applicability and effectiveness in diverse cinematic environments.

\section{Conclusion}
In this paper, we presents \textit{MovieCharacter}, a tuning-free framework for character video synthesis that effectively addresses the limitations of existing methods. By decomposing the synthesis process into distinct modules, our approach offers flexible customization while ensuring high-quality results. The integration of open-source models and established techniques further enhances the framework’s accessibility and efficiency, making it a valuable tool for a wide range of applications in film production, video game development, and interactive media. Future work may explore the expansion of \textit{MovieCharacter} to support even more complex interactions and broader character designs, thereby enriching the potential for creative expression in digital storytelling.



\bibliographystyle{ieee}
\bibliography{character_a_movie_arxiv}

\end{document}